\documentclass[10pt,twocolumn,letterpaper]{article}

\usepackage{iccv}
\usepackage{times}
\usepackage{epsfig}
\usepackage{graphicx}
\usepackage{amsmath}
\usepackage{amssymb}
\usepackage{multirow}
\usepackage{makecell}
\usepackage{booktabs}
\usepackage{caption}
\usepackage{subcaption}
\usepackage{float}

\usepackage[breaklinks=true,bookmarks=false]{hyperref}

\iccvfinalcopy 

\def\iccvPaperID{****} 

\ificcvfinal\fi

\begin{document}

\title{Mask Hierarchical Features For Self-Supervised Learning}

\author{Fenggang Liu\\
SenseTime Group\\
{\tt\small liufenggang@senseauto.com}
\and
Yangguang Li\\
SenseTime Group\\
{\tt\small liyangguang@sensetime.com}
\and
Feng Liang\\
University of Texas at Austin\\
{\tt\small jeffliang@utexas.edu}
\and
Jilan Xu\\
 Fudan University\\
{\tt\small 18210240039@fudan.edu.cn}
\and
Bin Huang\\
SenseTime Group\\
{\tt\small huangbin@senseauto.com}
\and
Jing Shao\\
SenseTime Group\\
{\tt\small shaojing@senseauto.com}
}



\maketitle
\ificcvfinal\thispagestyle{empty}\fi

\begin{abstract}
This paper shows that \textbf{Mask}ing the \textbf{Deep} hierarchical features is an efficient self-supervised method, denoted as \textbf{MaskDeep}. MaskDeep treats each patch in the representation space as an independent instance. We mask part of patches in the representation space and then utilize sparse visible patches to reconstruct high semantic image representation. 
The intuition of MaskDeep lies in the fact that models can reason from sparse visible patches semantic to the global semantic of the image. 
We further propose three designs in our framework: 1) a Hierarchical Deep-Masking module to concern the hierarchical property of patch representations, 2) a multi-group strategy to improve the efficiency without any extra computing consumption of the encoder and 3) a multi-target strategy to provide more description of the global semantic.
Our MaskDeep brings decent improvements. Trained on ResNet50 with 200 epochs, MaskDeep achieves state-of-the-art results of 71.2\% Top1 accuracy linear classification on ImageNet. On COCO object detection tasks, MaskDeep outperforms the self-supervised method SoCo, which specifically designed for object detection. When trained with 100 epochs, MaskDeep achieves 69.6\% Top1 accuracy, which surpasses current methods trained with 200 epochs, such as HCSC, by 0.4\% .
\end{abstract}

\section{Introduction}
\label{sec:intro}

In computer vision, self-supervised learning has attracted unprecedented attention for its impressive representation learning ability. Contrastive learning and Masked Image Modeling(MIM) are the two most effective methods. 

\begin{figure}[t]
     \centering   
    \includegraphics[width=\linewidth]{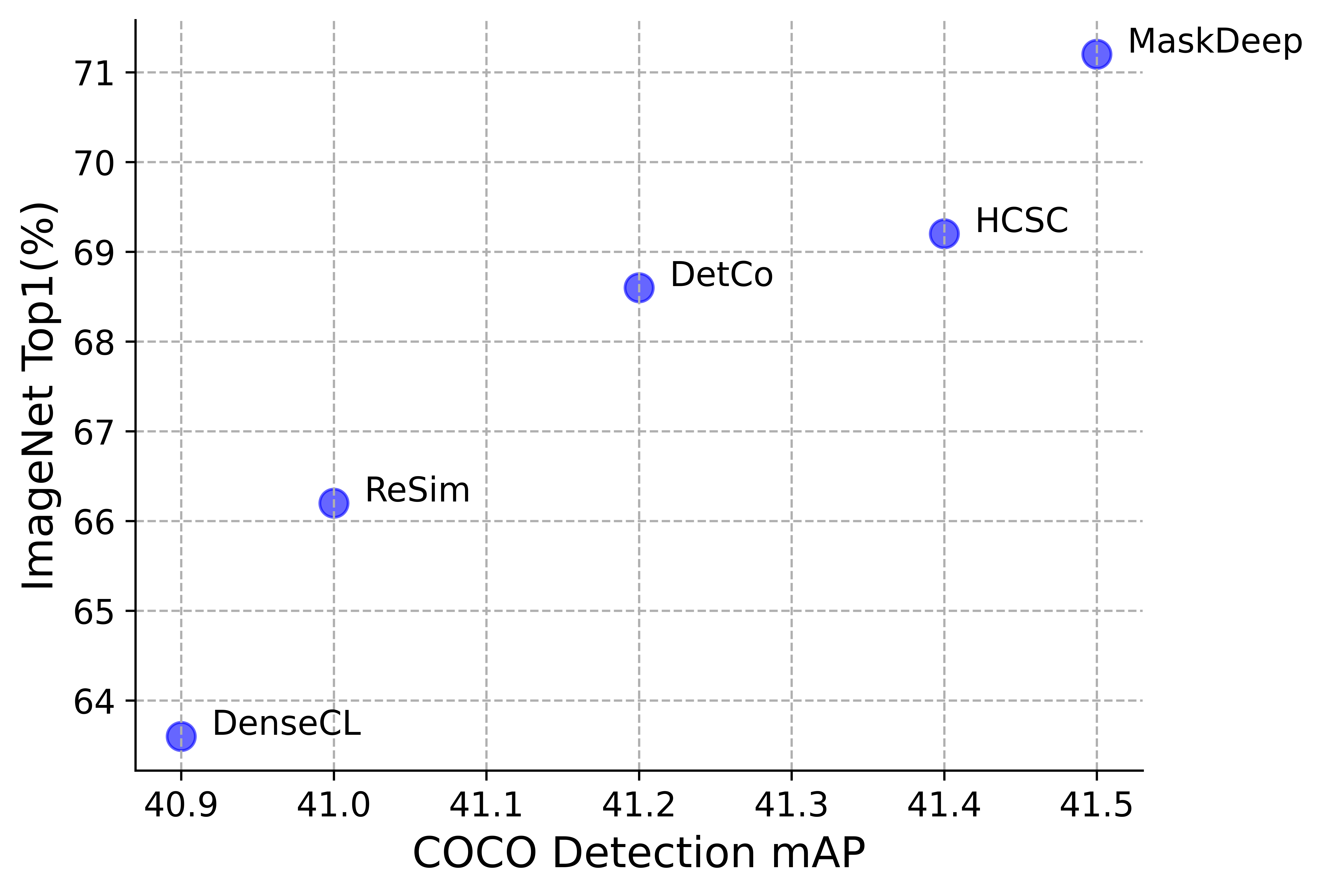}
     \caption{Compared with different contrastive learning methods on both ImageNet-1k classification and COCO detection benchmark. All methods are trained with 200 epoches on ImageNet-1k dataset.}
     \label{fig:performance}
\end{figure}

Contrastive learning methods are first adopted to learn invariant image-level global representation. Concretely, the network is typically trained to maximize the similarity between two augmentations of one image, such as MoCo v1/v2 \cite{he2020momentum,chen2020improved}, BYOL \cite{grill2020bootstrap}, and SimSiam \cite{simsiam}. These image-level methods can learn distinguished visual representation, such achieve high performance on classification tasks.
Other contrastive learning methods\cite{densecl,insloc,wei2021aligning} attempt to model local information by aligning the local representations from the same location in different augmented image views. For instance, SoCo\cite{wei2021aligning} first generated object proposals via selective search \cite{ren2015faster} and then aligned the pre-proposal object features of the different views. These object-level methods achieve high performance on dense prediction tasks but show a distinct disadvantage on classification tasks. 

Recently, Masked Image Modeling(MIM)\cite{he2022masked,bao2021beit,xie2022simmim,zhou2021ibot,baevski2022data2vec} methods have aroused great interest in the community. 
The core idea of MIM is to reconstruct the masked source image with sparse unmasked patches. Take MAE\cite{he2022masked} for example, it masks patches of the source images, feeds the masked image into an encoder model, and then utilizes visible patch features to predict the masked pixels by a decoder model. Other methods\cite{bao2021beit,xie2022simmim} concern the high-frequency noise of predicting low-level pixels. Therefore, they predict high-level semantics of the patches embedded by a tokenizer, instead of predicting low-level pixels directly. These MIM methods have made a breakthrough for the pre-training of the vision transformer model. We attribute part of the success of the MIM methods to the relationship probing among sparse patches. It proves that reasoning invisible context from sparse local patches is an efficient pretext task. 

In this paper, We follow the idea of learning knowledge from reasoning the relationship among sparse local patches. Distinctively, to avoid the consumption of thousands of training epochs, we extract the embedding of sparse local patches from a well-designed feature space several times, instead of feeding in masked images repeatedly, like MIM methods. Concerning the hierarchical property of visual representations, we incorporate an FPN-type module to make patch embedding hierarchical. We assume that each point feature in the spatial feature represents the high-level semantics of a specific region in the source image. Such, We can construct the group of sparse visible patch embedding form an implicit representation space directly. We called the approach as MaskDeep. 

\begin{figure*}[t]
     \centering   
    \includegraphics[width=0.7\linewidth]{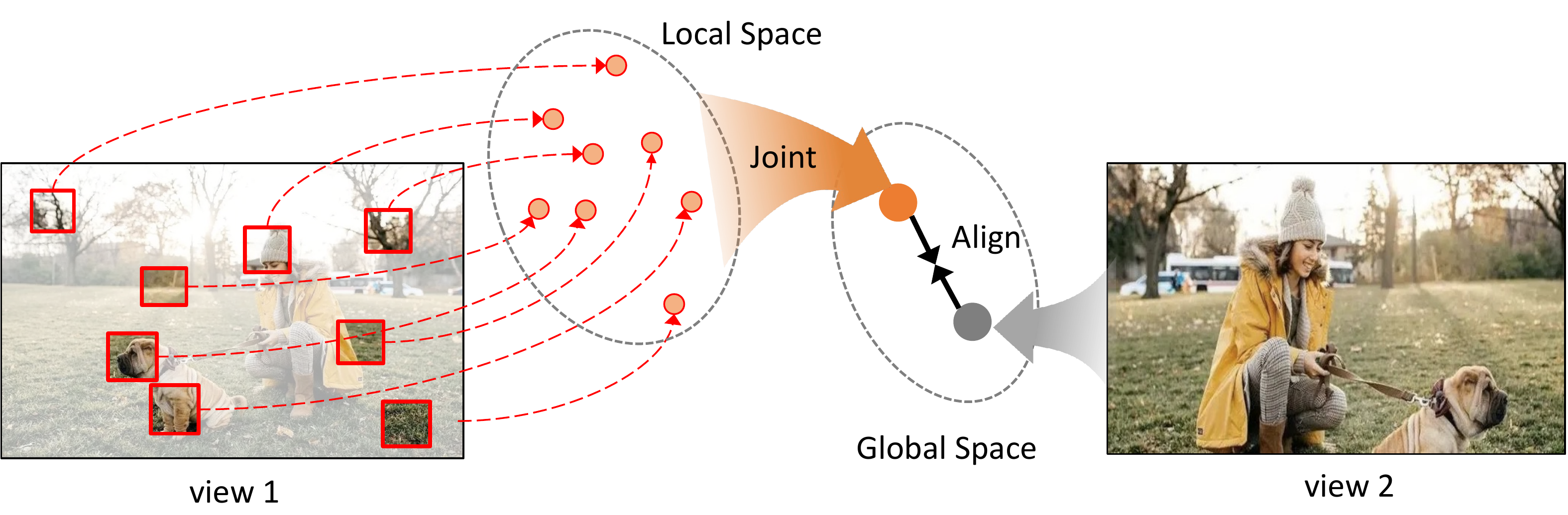}
     \caption{The supervision of MaskDeep. The local patches are embedded in the local feature space, while global images are embedded in the global feature space. Then, the local features are integrated to align with the global features.}
     \label{fig:motivation}
\end{figure*}

To learn knowledge from the groups of sparse local patches, we reason from sparse patch embedding to predict the global semantics of an image. The intuition lies in the fact that the local objects contain fine-grained object semantics, while the global image demonstrates high-level semantic information across all spatial locations. Integrating local object features to represent global image semantics can be an effective self-supervised objective. As the Figure~\ref{fig:motivation}, the sparse patch features, including 'human head', 'dog leg' and 'tree' \etc, are integrated to predict the image contents that 'A person is walking a dog in the park'.
In this way, both local object understanding and global context information reasoning are concerned. This property promises MaskDeep a unified self-supervised learner, which is friendly for both image-level and object-level downstream tasks. Concretely, in MaskDeep, an online branch is used to construct patch features and a momentum branch is used to extract the information of the whole image. Only about ten sparse patches are used to predict the information of the whole image. The prediction is supervised by maximizing the cosine similarity between the joint-patches representation and global representation. Different from the previous contrastive learning method~\cite{grill2020bootstrap}, MaskDeep needs no projector after the momentum encoder.

Extensive experiments on both classification and dense prediction downstream tasks show the effectiveness and generality of our MaskDeep. As the Figure~\ref{fig:performance}, trained on ResNet50 with 200 epochs, MaskDeep achieves 71.2\% Top1 accuracy linear classification on ImageNet, and 41.5\% bbox mAP on COCO object detection tasks, which surpasses current self-supervised methods. In the Table~\ref{tab:main}, MaskDeep outperforms the SoCo, which specifically designed for object detection, by 0.1 mAP. 

Our contributions can be summarized into three-folds. 
\begin{itemize}
    \item[$\bullet$] We propose a novel self-supervised learning approach, i.e., MaskDeep, in which views are masked from an well-designed implicit representation space.
    \item[$\bullet$] We align the joint-patches representation and the global representation directly without projector. Both local object understanding and global context information reasoning are concerned.
    \item[$\bullet$] MaskDeep achieves the state-of-the-art performance on both classification and dense prediction tasks, i.e., classification (71.2\% Top1 ACC), object detection (41.5\% box AP), and instance segmentation (36.1\% mask AP) when trained 200 epochs.
\end{itemize}

\section{Related Works}
\subsection{Contrastive Learning}
Contrastive-based self-supervised learning approaches \cite{he2020momentum,chen2020improved,grill2020bootstrap,simclr,instancediscri,xie2021detco,wei2021aligning,densecl,insloc,wang2022repre,li2021supervision,shao2021intern} have been proposed to circumvent the need for extensive annotations. The general purpose of these approaches is to learn useful image semantics via contrasting positive image pairs from negative image pairs. MoCo \cite{he2020momentum} proposed to maintain a memory queue of negative samples for contrastive loss, and thus relieved the need for large batch size during training. 
Another line of work only used positive pairs. For instance, BYOL \cite{grill2020bootstrap} proposes an asymmetric architecture, where a predictor is appended after the online branch and the target branch is updated via the momentum strategy. 

Beyond image-level self-supervised works, object-level self-supervised works \cite{xie2021detco,wei2021aligning,densecl,insloc} learn the image representation at the object-level, which mainly targets downstream tasks such as object detection and semantic segmentation. To name a few, InsLoc \cite{insloc} generated positive pairs in the instance-level contrastive loss by copying and pasting foreground instances to different background images. To maintain competitive performance on both classification task and instance-level dense prediction task, DetCo \cite{xie2021detco} generalized MoCo \cite{he2020momentum} to formulate instance-level supervision. DetCo obtained the joint local representation by aggregating the features of non-overlapping image patches. These works experimentally demonstrated the effectiveness of introducing properties of detection to self-supervised pretraining. 


\subsection{Masked Image Modeling}
Masked Image Modeling achieves outstanding performance in the field of self-supervised learning for transformer models.
BEiT \cite{bao2021beit} took the block-wise masked image and predicted the corresponding discrete visual tokens of the masked patches, where the target of the visual tokens are obtained directly from a pre-trained visual tokenizer. MAE \cite{he2022masked} exploited an autoencoder architecture to reconstruct the raw normalized RGB pixels of the masked patches, without the need of passing masked tokens into the encoder. Data2Vec \cite{baevski2022data2vec} and IBOT \cite{zhou2021ibot} proved the effectiveness of aligning the features of the masked image patches with the features of the whole image. 
Different from these approaches, our MaskDeep strategy performs masking over the deep features instead of the input image, and the goal is to align multiple deep, sparse features with the complete image features. 


\begin{figure*}[t]
     \centering
    \includegraphics[width=\linewidth]{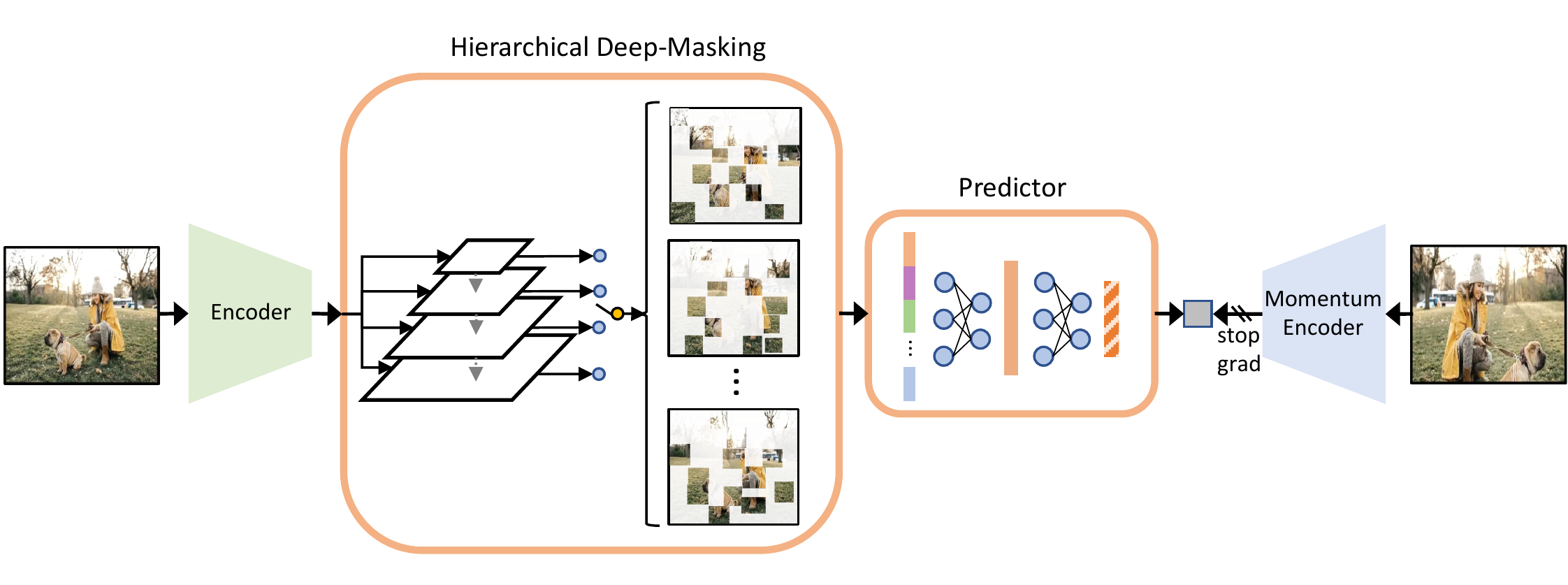}
     \caption{The framework of MaskDeep. MaskDeep uses sparse patch features from the hierarchical representation space to align the global semantics of the whole image. After feeding one view of image into the encoder, we generate the hierarchical representation space through the Hierarchical Deep-Masking module. We sample multi-group patches from the representation space. Then, each group of patches is decoded by a simple linear predictor. The global semantics of the whole image is generated by a momentum encoder from the other view of the image.}
     \label{fig:maskdeep}
\end{figure*}

\section{Method}
\label{sec:Framework}

\subsection{Overview}

The overall framework of our proposal self-supervised learning method is illustrated in Figure~\ref{fig:maskdeep}. MaskDeep follows the main paradigm of contrastive learning. A pair of encoder and momentum encoder are used to process two different views. The two views are preprocessed by augmentation methods of MoCov2\cite{chen2020improved}. MaskDeep focuses on probing the relationship of the local and the global semantic and reduces the computational overhead of building local semantic processes. We construct local semantic from an implicit representation space and integrate these local semantic to express the semantics of the whole image. The core design of MaskDeep lies in three parts: Hierarchical Deep-Masking Module, Predictor and Target, which will be described in details in the following section. 

\subsection{Hierarchical Deep-Masking Module}

Hierarchical Deep-Masking Module generates a hierarchical representation space, such make patch embedding containing hierarchical property. It is constructed based on the FPN module\cite{lin2017feature}. For ResNet, given an image, the encoder can generate image features, e.g. $\{C_2, C_3, C_4, C_5\}$, from different stages, where the feature $C_i$ corresponds to the $i_{th}$ stage. We feed the image features  $C_i$ into an FPN-type module. The outputs of the FPN-type module, e.g. $\{P_2, P_3, P_4, P_5\}$, are constructed as the hierarchical representation space. In this way, both low-level image features (e.g., $C_2$) and high-level image features (e.g., $C_5$) are considered to form a multi-scale representation space.  

Then, the outputs of the Hierarchical Deep-Masking module $\{P_2, P_3, P_4, P_5\}$ are randomly masked at the point level in the spatial dimension. On each scale output of the Hierarchical Deep-Masking module, we remain the same number $k$ of point features to ensure that each scale is given equal attention. From the remained $4\times k$ unmasked point features, we randomly sample $k$ point features at different scale level as the final group of patch features. In detail, each points is taken from each scale of the Hierarchical Deep-Masking Module module with equal probability. Such, the groups of patch features can remain the hierarchical property of visual representations.

Furthermore, we proposed a multi-group strategy, which is specialized in our method. Attributed to our deep-masking module, the group of patch features can be sampled several times without any extra computing consumption of the encoder. Each group of patch features is used to predict the global semantics of the image independently. The multi-group strategy increases the batch size in disguise, but consumes less computing resources.

\subsection{Predictor}
The predictor aggregates the groups of patch features by a simple two linear layers. The first linear layer is followed by a batch normalization\cite{ioffe2015batch}, and rectified linear units (ReLU)\cite{nair2010rectified}. This predictor has the same architecture with that in other contrastive learning methods. Denote a group of patch features as $\mathcal{H} = \{h_i\}_{i=1}^{k}$, where $h_i\in \mathrm{R}^d$ is a patch feature at one spatial location, $k$ is the number of patch feature in one group. In MaskDeep, we sample $K$ groups of patch features, which are denoted as $\{ \mathcal{H}_i \}_{i=1}^{K}$. Each group of patch features are concatenated to feed into the predictor. The predictor $q$ aggregates the group of patch representations $\mathcal{H}$ to a joint feature $q(\mathcal{H}_{i=1}^{K})$, which can describe the main content of the image. 

\subsection{Target}
Target is the direct output of the last layer from momentum encoder$f_\phi$, without any extra parameterized module such as the projector. For ResNet, the target is global pooled to a vector. All targets are detached from the computation graph. The target, denoted as $\mathcal{G}$, is expected to represent the global semantic of the image. 
The parameters of the momentum encoder are updated by the Exponentially Moving Average (EMA), according to the parameters of the online encoder. The EMA plays a critical role in preventing the global branch from model collapse \cite{he2020momentum,grill2020bootstrap}. The EMA is calculated as:
\begin{equation}
    \phi := \lambda\phi + (1 - \lambda)\theta,
\end{equation}
where $\lambda$ is the momentum coefficient, $\theta$ is the parameters of the online encoder, $\phi$ is the parameters of the momentum encoder. We scale up the coefficient from 0.99 to 1 during training. 

Further, we proposed a multi-target strategy. Different from the multi-crop strategy, in which multiple views are fed only into the online encoder, the multi-target strategy only feeds extra views into the momentum encoder. The momentum encoder is applied with the stop-gradient strategy, such no computational graph will be created. The increase of memory usage is small. In our multi-target strategy, a joint patch representation needs to predict multiple targets. The multi-target can be denoted as $\{ \mathcal{G}_j \}_{j=1}^{N}$, where $N$ is the number of target. The multi-target strategy provides more description of the global semantic, which can maintain the training stability .

\subsection{Loss Function}
The loss function is simply a cosine similarity loss, defined as $<\cdot, \cdot>$. The loss function aligns the joint patches feature representation $q(\mathcal{H})$ and the global representation $\mathcal{G}$. With the multi-group strategy, we can obtain $K$ joint patches features. With the multi-target strategy, we can obtain $N$ global features. Each pair of joint patches feature and global feature are aligned, as described in Eq.~\ref{eq:joint-mask}

\begin{equation}
    \begin{aligned}
    \mathcal{L} = -4 \cdot \sum_{i=1}^{K} \sum_{j=1}^{N} \ ( <p(\mathcal{H}^1_i),\mathcal{G}^2_j> + \\ <p(\mathcal{H}^2_i),\mathcal{G}^1_j>) ,
    \label{eq:joint-mask}
    \end{aligned}
\end{equation}
where the first term represents the alignment between mask representation of view 1 and the global representation of view 2. The second term swaps the views for the online and momentum branches. 

\begin{table*}[t]
\centering
\caption{The performances on different downstream tasks, including linear classification (CLS), object detection (DET) and instance segmentation (SEG). }
\resizebox{\linewidth}{!}{
\begin{tabular}{lc|cc|ccc|ccc|ccc}
\toprule
\multirow{2}{*}{Method} & \multirow{2}{*}{Epoch} & \multicolumn{2}{c|}{CLS} &
\multicolumn{3}{c|}{VOC-DET} & \multicolumn{6}{c}{COCO-DET-SEG 2x}  \\ \cline{3-13}
& & Top1& Top5&$\rm AP^{bb}$     &  $\rm AP^{bb}_{50}$ & $\rm AP^{bb}_{75}$ &   $\rm AP^{bb}$     &  $\rm AP^{bb}_{50}$ & $\rm AP^{bb}_{75}$ & $\rm AP^{mk}$     &  $\rm AP^{mk}_{50}$ & $\rm AP^{mk}_{75}$  \\
\midrule 
Supervised& 90 &  76.5 & - & 53.5  &  81.3 &  58.8  & 40.0     & 59.9 & 43.1  &  34.7      & 56.5 & 36.9  \\ 
MOCOv2   &  200  &  67.5 & - &  57.0    &  82.4 &  63.6 &  38.9      &  58.4 &  42.0 &  34.2    &  55.2&  36.5 \\ 
BYOL &  100 &  66.5  &  -       &   -   & -   & -&  -   & - &  -  &  -   & - &  - \\
BYOL  &  300 &  72.5   &  -       &   51.9   & 81.0   & 56.5 &  40.3     & 60.5 &  43.9  &  35.1    &  56.8 &  37.3 \\

DenseCL &  200 & 63.6 & 85.9 & 58.6&83.0&65.6  &  40.9 &  60.7 &  44.1 & 35.6 & 57.3 &  38.1    \\ 
ReSim-C4 &  200 &66.2 & 86.9 &  \textbf{58.6} & 83.0 & 65.6 & 41.0 & 60.7 & 44.5 & 35.7 & 57.6 & 38.0   \\ 
DetCo   &  200 &  68.6 &  88.5 &  57.8     & 82.6  &  64.2 &  41.2  &  61.0 &  44.7  &  35.8  &  57.8 &  38.2  \\ 

SoCo-C4  & 100 &59.7  &82.8 & {58.5} & \textbf{83.4} & \textbf{65.7} & 41.1 &  61.0&  44.4     & 35.6  &  57.5& 38.0  \\ 
    
HCSC   & 200 &  69.2 & - & - & 82.5 & - & 41.4& - &  -  & -  & - & -  \\ 

MaskDeep & 100 & 69.6 & 88.6 & 57.9 &  82.9 &  64.8 & 41.2 & 61.1 & 44.7 & 35.8 & 57.7 & 38.2    \\
MaskDeep & 200 &  \textbf{71.2} & \textbf{89.5} & 58.0& 82.8 & 64.7 & \textbf{41.5}& \textbf{61.1} & \textbf{45.3} & \textbf{36.1} & \textbf{58.0} & \textbf{38.8}   \\ 

\bottomrule
\end{tabular}
}

\label{tab:main}
\end{table*}

\section{Experiments}

To evaluate the generality of  MaskDeep, we evaluate  MaskDeep on both classification and dense prediction tasks, including ImageNet linear classification\cite{deng2009imagenet}, PASCAL VOC object detection\cite{everingham2010pascal}, COCO object detection and instance segmentation\cite{lin2014microsoft}. 

\subsection{Implementation Details}
\label{sec:imp}
Our  MaskDeep is pre-trained on ImageNet1k dataset, which consists of 1.28 million images. The data augmentation pipeline follows MoCov2.  We use the SGD optimizer with cosine decay learning rate schedule\cite{loshchilov2016sgdr} and a warm-up period of 10 epochs. The learning rate is set to 0.02. The batch size is 1024 when training 100 epochs. The weight decay is set to $10^{-5}$. The cosine decay momentum coefficient $\lambda$ starts from 0.99 and is increased to 1. When training 200 epochs, we increase the batch size to 2048 and double the learning rate to 0.04. 

We apply standard ResNet-50 as the backbone network with synchronized batch normalization\cite{ioffe2015batch} enabled. 
In the Hierarchical Deep-Masking module, we use the output of the stages $\{P_3, P_4, P_5\}$ for default. The output dimension of the Hierarchical Deep-Masking module is 512, which is the same with dimension of each patch representation. For the multi-group strategy, we sample 16 groups of patch features. For the multi-target strategy, we apply 4 extra global targets. Corresponding, in the Equation~\ref{eq:joint-mask}, K is 16, N is 4. In default, we sample 12 patch features in each group. The input dimension of predictor is equal to the sum of a group of patch features dimension. The hidden dimension of predictor is 4096. The output dimension of predictor is the same with that of the target, which is 2048. 

\subsection{Results}

We compare supervised and self-supervised learning methods with MaskDeep on a series of downstream tasks. 

\noindent\textbf{ImageNet linear classification.} Following MoCo \cite{he2020momentum}, we freeze the pre-trained backbone network and train a linear classification head for 100 epochs. 
The initial learning rate is 30 and divided by 10 at 60 and 80 epochs. Training images are randomly cropped and resized to 224$\times$224 and horizontally flipped with a probability of 0.5. 
As shown in Table~\ref{tab:main}, when trained with 200 epoches, MaskDeep achieves 71.2\% top-1 acc. and 89.5\% top-5 acc., significantly outperforming all self-supervised learning methods. When trained with 100 epoches, MaskDeep achieves 69.6\% top-1 acc, outperforming the HCSC trained with 200 epochs by 0.4\%. It indicates that MaskDeep is a more efficient representation learning method.

\noindent\textbf{PASCAL VOC object detection.} We adopt PASCAL VOC trainval07+12 for finetuning and report the results of $\text{AP}^{bb}$, $\text{AP}^{bb}_{50}$, and $\text{AP}^{bb}_{75}$ on test07 set. Following the protocol in  \cite{densecl,xie2021detco}, we finetune a Faster-RCNN object detector with C4-backbone. As shown in Table~\ref{tab:main}, MaskDeep is slightly inferior to the state-of-the-art method, which is specially designed for object detection task. MaskDeep shows a comparable results. 

\noindent\textbf{COCO object detection and instance segmentation.}
We compare the  MaskDeep with the state-of-the-art methods when transferring to object detection and instance segmentation on COCO 2017 dataset. We combine the pre-trained backbone network with a Mask RCNN detector (C4-backbone)\cite{he2017mask}. Following the settings in \cite{xie2021detco,wei2021aligning}, the standard 2$\times$ schedule is adopted. 
As shown in Table~\ref{tab:main}, under 200-epoch training,  MaskDeep obtains 41.5\% box AP and 36.1\% mask AP, which outperforms SoCo by 0.1\% and 0.2\%, respectively. For 200 epochs,  MaskDeep achieves slightly better results than other object-level self-supervised learning methods as well. 

In all, our MaskDeep brings decent improvements on both classification, object detection and segmentation tasks. MaskDeep shows an outstanding comprehensive ability. As obviously shown in the Figure~\ref{fig:performance}, MaskDeep surpasses current methods, including image-level and object-level self-supervised learning methods.



\section{Ablation Study}

We conduct a series of ablation studies step by step to demonstrate the effectiveness of each component in MaskDeep.
Unless specified, all the ablation experiments are constructed using a default batch size of 1024 and 100 pre-training epochs. The configuration for downstream tasks remains the same. The ResNet is evaluated on ImageNet linear-probing benchmark.

\begin{table}[h]
	\centering
	\caption{Ablation studies results on ImageNet linear-prob benchmark.}
        \resizebox{\linewidth}{!}{
	\begin{tabular}{c|ccc|c}
        \toprule  
        \makecell{Model} &
        \makecell{Hierarchical \\ Module}&
        \makecell{Multi-Group} &
        \makecell{Multi-Target} &
        Top1 \\ 
        \midrule   
        BYOL    &           &            &            & 66.5   \\ 
        Naive   &           &            &            & 64.9   \\ 
        MaskDeep &\checkmark &            &             & 67.9  \\ 
        MaskDeep &\checkmark & \checkmark &             & 68.8  \\ 
        MaskDeep &\checkmark & \checkmark &  \checkmark & \textbf{69.6}    \\ 
        \bottomrule
	\end{tabular}
 }
\label{tab:ablation}
\end{table}

Because we follow the main paradigm of contrastive learning methods and use the cosine similarity loss function, we regard the BYOL as a baseline. 
The ablation results are shown in the Table~\ref{tab:ablation}. The first line represents the performance of the BYOL when trained 100 epoches. The second line represents a naive idea for masking patches in the representation space. In the naive idea, the group of point features are  sampled from the last layer of the encoder directly. As shown in the Table~\ref{tab:ablation}, the naive idea achieves 64.9\% top-1 acc, while SimCLR achieves 64.8\% and BYOL achieves 66.5\% with the same 100 training epochs. The baseline shows a little poor performance, but it works and learns representation from the naive pretext task. It lags behind image-level self-supervised methods, but surpasses object-level self-supervised methods, such as DenseCL which only achieve 63.6\%, as shown in the Table~\ref{tab:main}.

After concerning hierarchical property by the Hierarchical Deep-Masking module, our MaskDeep obtains an improvement of +3.0\%. It achieves 67.9\% top-1 acc and surpasses the BYOL by 1.4\%. With the Hierarchical Deep-Masking module, it is comparable with the current self-supervised learning methods.
Further, we introduce a multi-group strategy, which is specific to our method. Intuitively, the multi-group strategy increase the number of samples, such  obtains an improvement of +0.9\%. 
Finally, we introduce a multi-target strategy to provide more supervision signals. The strategy makes the training process more stable. As shown in the Table~\ref{tab:ablation}, an improvement of +0.8\% for ResNet is found. 

\begin{figure*}[t]
     \centering
    \includegraphics[width=\linewidth]{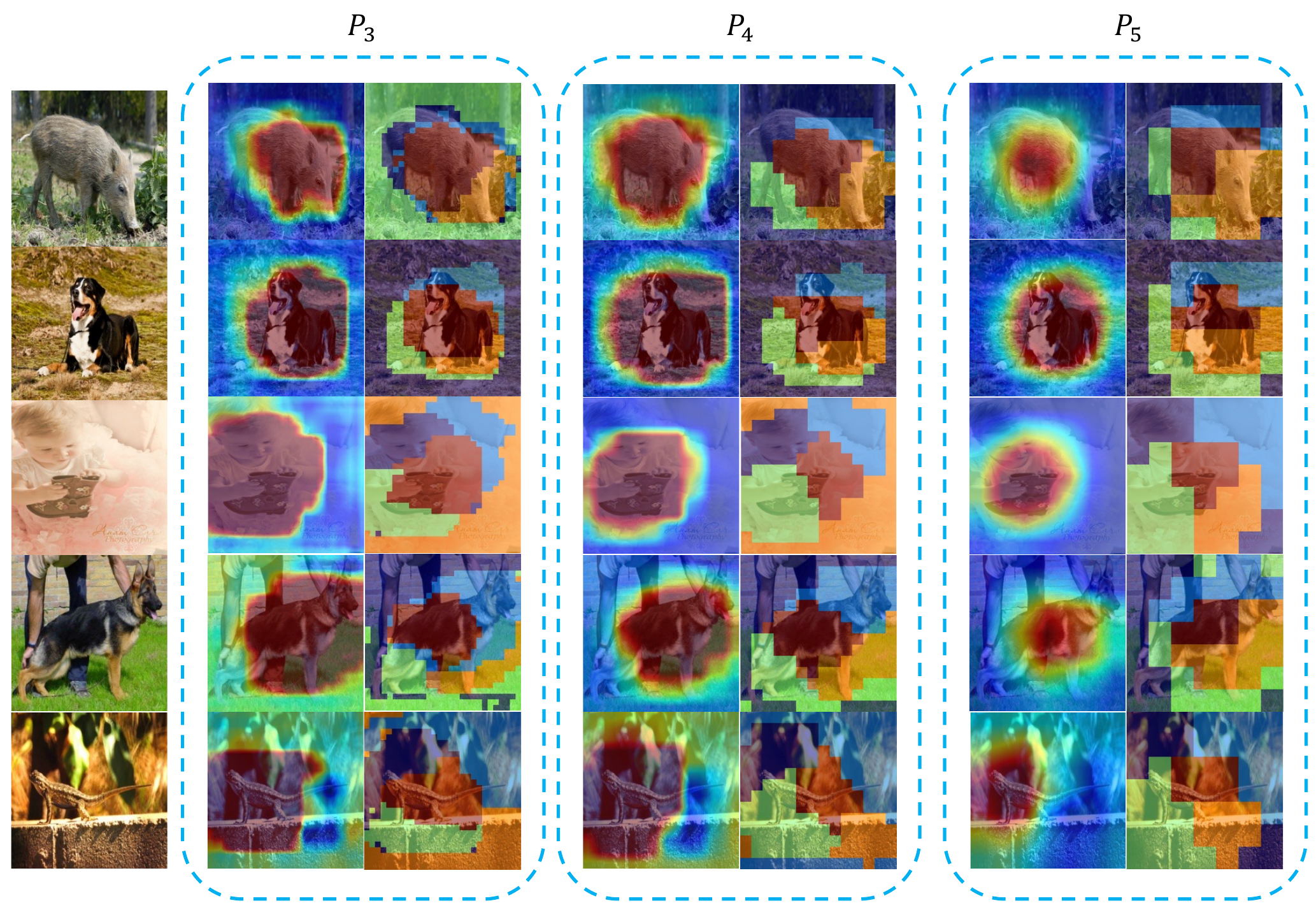}
     \caption{The visualization of the implicit representation space. We plot the e results of each scale level $\{P_3, P_4, P_5\}$ in the Hierarchical Deep-Masking module. For each scale level, the first column is the Grad-CAM results, the second column is the clustering map results.}
     \label{fig:visual}
\end{figure*}

\section{Analysis}

In this section, we further examine different hyperparameters for each component in MaskDeep. 

\noindent\textbf{The scale level and dimension of the Hierarchical Deep-Masking module.}
The Table~\ref{table:hierarchical} shows the influence of the number of scale level and dimension for the Hierarchical Deep-Masking module. To understand the influence of scales, we ablate the layer number of Hierarchical Deep-Masking module from the min scale to max scale. 

In the Table~\ref{table:hierarchical} (a), the level 1 represents only using the output of the last stage $\{P_5\}$, the level 2 represents using the output of the last two stages $\{P_4, P_5\}$, the level 3 represents using the output of the stages $\{P_3, P_4, P_5\}$, the level 4 represents using the output of the all stages $\{P_2, P_3, P_4, P_5\}$,. The results show that when only using the output of the last stage $\{P_5\}$, the performance drops about 1.4\%. It indicates the indispensable role of the Hierarchical Deep-Masking module. Except only using the $\{P_5\}$, our method is robust to the other levels. The results show that level 4 is a little harmful. This could be caused by the large resolution of $\{P_2\}$, the semantic information of each point is insufficient. Our method achieves the best performance with the output of the stages $\{P_3, P_4, P_5\}$. 

\begin{table}[h]
    \centering
    \caption{The analysis for the scale level and dimension of the Hierarchical Deep-Masking module.}
    \label{table:hierarchical}
    \subfloat[Scale Level.]{
        \begin{tabular}{ccccc}
        \toprule
        Level   &  1 & 2 & 3 & 4 \\
        \midrule 
        Top1 & 67.4 & 68.7 & \textbf{68.8}  & 68.6 \\
        \bottomrule
        \end{tabular}
	}
    \\
    \subfloat[Dimension.]{
        \begin{tabular}{ccccc}
        \toprule
        Dim   & 256 & 512 & 1024 \\
        \midrule
        Top1 & 68.5 & \textbf{68.8} & 68.6 \\
        \bottomrule
        \end{tabular}
	}
\end{table}

Table~\ref{table:hierarchical} (b) shows the influence of the dimension of the Hierarchical Deep-Masking module. While the output dimension of ResNet50 is $[256,512,1024,2048]$, the dimension 512 performs best. We summarize that the dimension should be close to the output dimension of the encoder. 
Small dimension will limit the representation capacity of each patch feature, while large dimension will decrease the training speed and increase learning difficulty.

\noindent\textbf{The number of patch features in one group.} We analyze the number of visible patch features. The results of different number of patch features is shown in Table~\ref{tab:patch}. MaskDeep is found to be insensitive to the patch number from 8 to 20. MaskDeep achieves best performance when the number is 12. Too little patches can not adequately cover the content of the image, while too many patches introduce repeated patch representation and noise, which increases learning difficulty.

\begin{table}[h]
\centering
\caption{The analysis for the number of patch features in each group.}
\begin{tabular}{ccccccc}
\toprule
Patch Num  & 4 & 8  & 12 & 16   & 20 & 24 \\
\midrule 
Top1 & 68.4  & 68.7 & \textbf{68.8} & 68.7 & 68.8 & 68.6\\
\bottomrule
\end{tabular}
\label{tab:patch}
\end{table}

\noindent\textbf{The number of groups in the multi-group strategy.} The Table~\ref{tab:multigroup} shows the impact of group number. When the group number is 1, representing that only one group of patches are sampled, the multi-group strategy is not used. When the group number increased from 1 to 8, our method obtains a great improvement of +0.6\%. When the group number increased from 8 to 16, the performance is improved saturatly, and only increased by 0.3\%. When we continue to increase the group number to 24, the performance is no longer improved, and even the performance has slightly decreased. 

\begin{table}[h]
\centering
\caption{The analysis for the number of groups in the multi-group strategy.}
\begin{tabular}{cccccc}
\toprule
Group Num  & 1   & 8   & 16  & 24   \\
\midrule 
Top1 &  67.9  & 68.5 &  \textbf{68.8}    &  68.7 \\
\bottomrule
\end{tabular}

\label{tab:multigroup}
\end{table}

\noindent\textbf{The number of targets in the multi-target strategy.} The Table~\ref{tab:multitarget} shows that the performance gets better as the target number increases from 0 to 4. Compared with using no multi-target strategy, 4 extra targets can improve performance by 0.8\%. But more targets benefit little. We think the multi-target strategy can provide a stable supervisory signal.

\begin{table}[h]
\centering
\caption{The analysis for the number of targets in the multi-target number.}
\begin{tabular}{cccccc}
\toprule
Target Num  & 0  & 2  & 4 & 6 \\
\midrule 
Top1 &  68.8  &69.0  &\textbf{69.6}  &  69.5  \\
\bottomrule
\end{tabular}

\label{tab:multitarget}
\end{table}

\noindent\textbf{The hidden dimension of the predictor.} The predictor contains two linear layers. As previous works show, the hidden dimension of the predictor is an important hyper-parameter. In this part, we examine the hyper-parameter as the Table~\ref{tab:predim}. The results show that the dimension 4096 performs best, while the patch number is 12 and the patch dimension is 512.

\begin{table}[h]
\centering
\caption{The analysis for the hidden dimension of the predictor.}
\begin{tabular}{cccc}
\toprule
Dim  & 2048  & 4096  & 8192 \\
\midrule 
 Top1 & 68.5 &   \textbf{68.8}  &   68.7     \\
\bottomrule
\end{tabular}
\label{tab:predim}
\label{tab:main}
\end{table}

\noindent\textbf{With the multi-crop strategy.} The multi-crop strategy~\cite{swav} is a powerful trick in self-supervised leaning methods. Previous works confirm that more views can improve the effect of self-supervised learning models. We apply the multi-crop strategy to our MaskDeep. We use a multi-scale multi-crop strategy as HSCS~\cite{guo2022hcsc}. In detail, three different scale views $\{192,160,128\}$ are used and each scale is sampled twice.
The results are shown in the Table~\ref{tab:mc}. With the multi-crop strategy, MaskDeep achieves a 2.5\% performance increase when training 100 epoches and a 2.8\% performance increase when training 200 epoches. After adding multi-crop strategy, MaskDeep also outperforms previous methods.

\begin{table}[h]
\centering
\caption{The performance with the multi-crop strategy. All methods are trained with the multi-crop strategy.}
\begin{tabular}{lcc}
\toprule
Method  & Epoch & Top1 \\
\midrule 
UniVIP & 200 & 73.1 \\ 
HCSC   & 200 & 73.3 \\ 
VICRegL\cite{bardes2022vicregl}  & 300 &  71.2 \\
MaskDeep  & 100 & 73.1 \\
MaskDeep  & 200 & \textbf{74.0} \\
\bottomrule
\end{tabular}
\label{tab:mc}
\end{table}

\section{Visualization}

The implicit representation space from the Hierarchical Deep-Masking module is an important part in our method. We sample all patch features from the implicit representation space. We assume that each patch features contains local semantic corresponding to specific image region, such we can infer from several patch features to the global image content. Therefore, we visualize the implicit representation space to verify our idea with Grad-CAM~\cite{selvaraju2017grad} and clustering map. In the visualization process, we only apply the resize preprocessing method for view1. When applying Grad-CAM, we only use one local representation $q(\mathcal{H}^1)$ and one global representation $\mathcal{G}^2$ to calculate the cosine similarity loss. We use the gradient at each scale level as the weight for the output of the Hierarchical Deep-Masking module. For clustering map, we select 5 point features as the initial features. Then, we calculate the similarity between all point features across space dimension with the 5 initial features. Finally, we cluster the points with highest similarity and show a cluster with the same color.

Fig. \ref{fig:visual} demonstrates the visualization results.
The Grad-CAM results show that all scale level of the Hierarchical Deep-Masking module can attention the main object in the image successfully. Regions covering the objects have higher activation, while background regions (e.g., ground or sky) receive lower attention. For different scale level, the granularity of attention is different. $P_3$, with largest resolution, shows the fine-grain attention. In the first image, the contours of pig head are accurately perceived.
The clustering maps show that patch features with similar context can be clustered together. The background and the subject are distinguishable. The distinguish is shown more clearly in fine-grain $P_3$ level. Take the first pig picture for example, the background is clustered into the same cluster and the most pig body have high semantic similarity.
Further, among different scale level, the clustering results are consistent. The visualization results support our assumption that each patch features contains specific local semantic. The process of aligning implicit local patch representation and global image representation is meaningful.

\section{Conclusions}
In this paper, we propose a simple but strong self-supervised learning method, i.e., MaskDeep. We aligns local and global representations in a single online-momentum framework. We show that reasoning from sparse patch features to infer the global image content is a strong self-supervised learning task. Furthermore, we first construct the patch features from an implicit representation space instead of masking the source image. We visualize the implicit representation space. The results prove that each point contains the semantic information of the corresponding context. We propose three novel designs: the hierarchical deep-masking module, the multi-group strategy and the multi-target strategy. These designs improve the efficiency of our method. Our MaskDeep outperforms current self-supervised methods on classification, object detection and segmentation downstream tasks. Trained on ResNet50 with 200 epochs, MaskDeep achieves state-of-the-art results of 71.2\% top-1 accuracy linear classification on the ImageNet dataset. On COCO object detection tasks, MaskDeep achieves state-of-the-art results of 41.5\% bbox mAP and 36.1 mask mAP, which outperforms most object-level self-supervised method. 

{\small
\bibliographystyle{ieee_fullname}
\bibliography{egbib}

\begin{thebibliography}{10}\itemsep=-1pt

\bibitem{baevski2022data2vec}
Alexei Baevski, Wei-Ning Hsu, Qiantong Xu, Arun Babu, Jiatao Gu, and Michael
  Auli.
\newblock Data2vec: A general framework for self-supervised learning in speech,
  vision and language.
\newblock In {\em International Conference on Machine Learning}, pages
  1298--1312. PMLR, 2022.

\bibitem{bao2021beit}
Hangbo Bao, Li Dong, Songhao Piao, and Furu Wei.
\newblock Beit: Bert pre-training of image transformers.
\newblock {\em arXiv preprint arXiv:2106.08254}, 2021.

\bibitem{bardes2022vicregl}
Adrien Bardes, Jean Ponce, and Yann LeCun.
\newblock Vicregl: Self-supervised learning of local visual features.
\newblock {\em arXiv preprint arXiv:2210.01571}, 2022.

\bibitem{swav}
Mathilde Caron, Ishan Misra, Julien Mairal, Priya Goyal, Piotr Bojanowski, and
  Armand Joulin.
\newblock Unsupervised learning of visual features by contrasting cluster
  assignments.
\newblock {\em Advances in Neural Information Processing Systems},
  33:9912--9924, 2020.

\bibitem{simclr}
Ting Chen, Simon Kornblith, Mohammad Norouzi, and Geoffrey Hinton.
\newblock A simple framework for contrastive learning of visual
  representations.
\newblock In {\em International conference on machine learning}, pages
  1597--1607. PMLR, 2020.

\bibitem{chen2020improved}
Xinlei Chen, Haoqi Fan, Ross Girshick, and Kaiming He.
\newblock Improved baselines with momentum contrastive learning.
\newblock {\em arXiv preprint arXiv:2003.04297}, 2020.

\bibitem{simsiam}
Xinlei Chen and Kaiming He.
\newblock Exploring simple siamese representation learning.
\newblock In {\em Proceedings of the IEEE/CVF Conference on Computer Vision and
  Pattern Recognition}, pages 15750--15758, 2021.

\bibitem{deng2009imagenet}
Jia Deng, Wei Dong, Richard Socher, Li-Jia Li, Kai Li, and Li Fei-Fei.
\newblock Imagenet: A large-scale hierarchical image database.
\newblock In {\em 2009 IEEE conference on computer vision and pattern
  recognition}, pages 248--255. Ieee, 2009.

\bibitem{everingham2010pascal}
Mark Everingham, Luc Van~Gool, Christopher~KI Williams, John Winn, and Andrew
  Zisserman.
\newblock The pascal visual object classes (voc) challenge.
\newblock {\em International journal of computer vision}, 88(2):303--338, 2010.

\bibitem{grill2020bootstrap}
Jean-Bastien Grill, Florian Strub, Florent Altch{\'e}, Corentin Tallec, Pierre
  Richemond, Elena Buchatskaya, Carl Doersch, Bernardo Avila~Pires, Zhaohan
  Guo, Mohammad Gheshlaghi~Azar, et~al.
\newblock Bootstrap your own latent-a new approach to self-supervised learning.
\newblock {\em Advances in Neural Information Processing Systems},
  33:21271--21284, 2020.

\bibitem{guo2022hcsc}
Yuanfan Guo, Minghao Xu, Jiawen Li, Bingbing Ni, Xuanyu Zhu, Zhenbang Sun, and
  Yi Xu.
\newblock Hcsc: hierarchical contrastive selective coding.
\newblock In {\em Proceedings of the IEEE/CVF Conference on Computer Vision and
  Pattern Recognition}, pages 9706--9715, 2022.

\bibitem{he2022masked}
Kaiming He, Xinlei Chen, Saining Xie, Yanghao Li, Piotr Doll{\'a}r, and Ross
  Girshick.
\newblock Masked autoencoders are scalable vision learners.
\newblock In {\em Proceedings of the IEEE/CVF Conference on Computer Vision and
  Pattern Recognition}, pages 16000--16009, 2022.

\bibitem{he2020momentum}
Kaiming He, Haoqi Fan, Yuxin Wu, Saining Xie, and Ross Girshick.
\newblock Momentum contrast for unsupervised visual representation learning.
\newblock In {\em Proceedings of the IEEE/CVF conference on computer vision and
  pattern recognition}, pages 9729--9738, 2020.

\bibitem{he2017mask}
Kaiming He, Georgia Gkioxari, Piotr Doll{\'a}r, and Ross Girshick.
\newblock Mask r-cnn.
\newblock In {\em Proceedings of the IEEE international conference on computer
  vision}, pages 2961--2969, 2017.

\bibitem{ioffe2015batch}
Sergey Ioffe and Christian Szegedy.
\newblock Batch normalization: Accelerating deep network training by reducing
  internal covariate shift.
\newblock In {\em International conference on machine learning}, pages
  448--456. PMLR, 2015.

\bibitem{li2021supervision}
Yangguang Li, Feng Liang, Lichen Zhao, Yufeng Cui, Wanli Ouyang, Jing Shao,
  Fengwei Yu, and Junjie Yan.
\newblock Supervision exists everywhere: A data efficient contrastive
  language-image pre-training paradigm.
\newblock {\em arXiv preprint arXiv:2110.05208}, 2021.

\bibitem{lin2017feature}
Tsung-Yi Lin, Piotr Doll{\'a}r, Ross Girshick, Kaiming He, Bharath Hariharan,
  and Serge Belongie.
\newblock Feature pyramid networks for object detection.
\newblock In {\em Proceedings of the IEEE conference on computer vision and
  pattern recognition}, pages 2117--2125, 2017.

\bibitem{lin2014microsoft}
Tsung-Yi Lin, Michael Maire, Serge Belongie, James Hays, Pietro Perona, Deva
  Ramanan, Piotr Doll{\'a}r, and C~Lawrence Zitnick.
\newblock Microsoft coco: Common objects in context.
\newblock In {\em European conference on computer vision}, pages 740--755.
  Springer, 2014.

\bibitem{loshchilov2016sgdr}
Ilya Loshchilov and Frank Hutter.
\newblock Sgdr: Stochastic gradient descent with warm restarts.
\newblock {\em arXiv preprint arXiv:1608.03983}, 2016.

\bibitem{nair2010rectified}
Vinod Nair and Geoffrey~E Hinton.
\newblock Rectified linear units improve restricted boltzmann machines.
\newblock In {\em Icml}, 2010.

\bibitem{ren2015faster}
Shaoqing Ren, Kaiming He, Ross Girshick, and Jian Sun.
\newblock Faster r-cnn: Towards real-time object detection with region proposal
  networks.
\newblock {\em Advances in neural information processing systems}, 28:91--99,
  2015.

\bibitem{selvaraju2017grad}
Ramprasaath~R Selvaraju, Michael Cogswell, Abhishek Das, Ramakrishna Vedantam,
  Devi Parikh, and Dhruv Batra.
\newblock Grad-cam: Visual explanations from deep networks via gradient-based
  localization.
\newblock In {\em Proceedings of the IEEE international conference on computer
  vision}, pages 618--626, 2017.

\bibitem{shao2021intern}
Jing Shao, Siyu Chen, Yangguang Li, Kun Wang, Zhenfei Yin, Yinan He, Jianing
  Teng, Qinghong Sun, Mengya Gao, Jihao Liu, et~al.
\newblock Intern: A new learning paradigm towards general vision.
\newblock {\em arXiv preprint arXiv:2111.08687}, 2021.

\bibitem{wang2022repre}
Luya Wang, Feng Liang, Yangguang Li, Wanli Ouyang, Honggang Zhang, and Jing
  Shao.
\newblock Repre: Improving self-supervised vision transformer with
  reconstructive pre-training.
\newblock {\em arXiv preprint arXiv:2201.06857}, 2022.

\bibitem{densecl}
Xinlong Wang, Rufeng Zhang, Chunhua Shen, Tao Kong, and Lei Li.
\newblock Dense contrastive learning for self-supervised visual pre-training.
\newblock In {\em Proceedings of the IEEE/CVF Conference on Computer Vision and
  Pattern Recognition}, pages 3024--3033, 2021.

\bibitem{wei2021aligning}
Fangyun Wei, Yue Gao, Zhirong Wu, Han Hu, and Stephen Lin.
\newblock Aligning pretraining for detection via object-level contrastive
  learning.
\newblock {\em Advances in Neural Information Processing Systems}, 34, 2021.

\bibitem{instancediscri}
Zhirong Wu, Yuanjun Xiong, Stella~X Yu, and Dahua Lin.
\newblock Unsupervised feature learning via non-parametric instance
  discrimination.
\newblock In {\em Proceedings of the IEEE conference on computer vision and
  pattern recognition}, pages 3733--3742, 2018.

\bibitem{xie2021detco}
Enze Xie, Jian Ding, Wenhai Wang, Xiaohang Zhan, Hang Xu, Peize Sun, Zhenguo
  Li, and Ping Luo.
\newblock Detco: Unsupervised contrastive learning for object detection.
\newblock In {\em Proceedings of the IEEE/CVF International Conference on
  Computer Vision}, pages 8392--8401, 2021.

\bibitem{xie2022simmim}
Zhenda Xie, Zheng Zhang, Yue Cao, Yutong Lin, Jianmin Bao, Zhuliang Yao, Qi
  Dai, and Han Hu.
\newblock Simmim: A simple framework for masked image modeling.
\newblock In {\em Proceedings of the IEEE/CVF Conference on Computer Vision and
  Pattern Recognition}, pages 9653--9663, 2022.

\bibitem{insloc}
Ceyuan Yang, Zhirong Wu, Bolei Zhou, and Stephen Lin.
\newblock Instance localization for self-supervised detection pretraining.
\newblock In {\em Proceedings of the IEEE/CVF Conference on Computer Vision and
  Pattern Recognition}, pages 3987--3996, 2021.

\bibitem{zhou2021ibot}
Jinghao Zhou, Chen Wei, Huiyu Wang, Wei Shen, Cihang Xie, Alan Yuille, and Tao
  Kong.
\newblock ibot: Image bert pre-training with online tokenizer.
\newblock {\em arXiv preprint arXiv:2111.07832}, 2021.

\end{thebibliography}
}

\end{document}